\documentclass[letterpaper]{article} 
\usepackage{aaai2026}  
\usepackage{times}  
\usepackage{helvet}  
\usepackage{courier}  
\usepackage[hyphens]{url}  
\usepackage{graphicx} 

\usepackage{tabularx}
\usepackage{enumitem}
\usepackage{amsmath}
\usepackage{amssymb}
\usepackage{siunitx}
\usepackage{multirow}
\usepackage{booktabs}
\usepackage{makecell}

\usepackage{times}
\usepackage{helvet}
\usepackage{courier}
\usepackage{xcolor}

\usepackage{natbib}  
\usepackage{caption} 
\usepackage{algorithm}
\usepackage{algorithmic}

\usepackage{newfloat}
\usepackage{listings}
\DeclareCaptionStyle{ruled}{labelfont=normalfont,labelsep=colon,strut=off} 
\floatstyle{ruled}
\newfloat{listing}{tb}{lst}{}
\floatname{listing}{Listing}
\title{Understanding the Essence: Delving into Annotator\\ Prototype Learning for Multi-Class Annotation Aggregation}
\author{
    Ju Chen\textsuperscript{\rm 1, \rm 2}, Jun Feng\textsuperscript{\rm 1, \rm 2}, Shenyu Zhang\textsuperscript{\rm 3, \rm 4, \rm 5}\\
}
\affiliations{
    \textsuperscript{\rm 1}Key Laboratory of Water Big Data Technology of Ministry of Water Resources, Hohai University, Nanjing, China\\

    \textsuperscript{\rm 2}College of Computer Science and Software Engineering, Hohai University, Nanjing, China\\
    \textsuperscript{\rm 3}School of Computer Science and Engineering, Southeast University, Nanjing, China\\
    \textsuperscript{\rm 4}Key Laboratory of New Generation Artificial Intelligence Technology and Its Interdisciplinary
Applications (Southeast University), Ministry of Education, China\\
\textsuperscript{\rm 5}Faculty of Information Technology, Monash University, Melbourne, Australia\\

    \{JuJuChen, fengjun\}@hhu.edu.cn, shenyuzhang@seu.edu.cn
}

\usepackage{bibentry}

\begin{document}

\maketitle

\begin{abstract}
Multi-class classification annotations have significantly advanced AI applications, with truth inference serving as a critical technique for aggregating noisy and biased annotations. Existing state-of-the-art methods typically model each annotator's expertise using a confusion matrix. However, these methods suffer from two widely recognized issues: 1) when most annotators label only a few tasks, or when classes are imbalanced, the estimated confusion matrices are unreliable, and 2) a single confusion matrix often remains inadequate for capturing each annotator's full expertise patterns across all tasks. To address these issues, we propose a novel confusion-matrix-based method, \textbf{PTBCC} (\textbf{P}roto\textbf{T}ype learning-driven \textbf{B}ayesian \textbf{C}lassifier \textbf{C}ombination), to introduce a reliable and richer annotator estimation by prototype learning. Specifically, we assume that there exists a set $S$ of prototype confusion matrices, which capture the inherent expertise patterns of all annotators. Rather than a single confusion matrix, the expertise per annotator is extended as a Dirichlet prior distribution over these prototypes. This prototype learning-driven mechanism circumvents the data sparsity and class imbalance issues, ensuring a richer and more flexible characterization of annotators. Extensive experiments on 11 real-world datasets demonstrate that PTBCC achieves up to a 15\% accuracy improvement in the best case, and a 3\% higher average accuracy while reducing computational cost by over 90\%.
\end{abstract}


\section{Introduction}
Multi-class crowdsourced annotations, which advance AI applications by producing tremendously valuable annotations and facilitates fact-checking \cite{Li24,Otanietal23,Chenetal22}, are increasingly important with the rise of large language models \cite{yao2024a,Gienappetal25}. Given the varied annotator expertise, the collected annotations are often noisy and biased \cite{HubeFG19,Lietal2019a}. Therefore, crowdsourcing platforms typically assign each task to multiple annotators \cite{Sheng19,Wallaceetal22}. The most naive truth inference method, Majority Voting (MV), treats the annotation provided by most annotators as the truth \cite{Li14}. While easy to implement, it ignores differences in annotator expertise and is often suboptimal in practice \cite{Hornetal18}. To obtain accurate annotations, a wide range of truth inference methods have been proposed to iteratively infer annotator expertise and ground truth.

Existing methods can be grouped into single-parameter-based approaches and confusion-matrix-based approaches, depending on whether they distinguish an annotator's classification expertise under different truths \cite{Chenetal25}. Among single-parameter-based approaches, BWA \cite{Lietal2019b} demonstrates the best generalization ability by modeling each annotator as drawn from a Gamma distribution, and maintains efficiency second only to MV. However, owing to the extensive representation of annotators, confusion-matrix-based methods generally perform better and show greater promise \cite{Zhangetal24}. \citet{DawidSkene1979} are the first to model each annotator's expertise using a confusion matrix. \citet{KimGhahramani2012} extend \cite{DawidSkene1979} by incorporating Dirichlet priors. \citet{Lietal2019a} classify tasks with the same truth into subtypes to capture fine-grained annotator expertise. \citet{Zhangetal24} propose a coupled confusion correction method to mutually refine the learned confusion matrices. \citet{Chenetal25} develop a Softmax-Gaussian structure to replace the Dirichlet-Multinomial pattern of \citet{KimGhahramani2012}.

However, multi-class classification annotations pose significantly more challenges than binary classification ones \cite{Chu21a,Yangetal24,Zhangetal25}, and existing confusion-matrix-based methods suffer from the following widely acknowledged issues: 1) when most annotators label only a few tasks or when classes are imbalanced, the estimated confusion matrices are unreliable due to data sparsity and class imbalance, and 2) a single confusion matrix is often inadequate to capture each annotator's complex expertise patterns across all tasks. Moreover, current confusion-matrix-based methods often struggle with high computational overhead, which hinders their scalability.

To address the above issues, we propose a novel confusion-matrix-based method, \textbf{PTBCC} (\textbf{P}roto\textbf{T}ype learning-driven \textbf{B}ayesian \textbf{C}lassifier \textbf{C}ombination). Specifically, we assume that there exists a set $S$ of prototype confusion matrices, which capture the inherent expertise patterns of all annotators, and the expertise of each annotator follows a Dirichlet prior distribution over these prototypes. By constructing $|S|$ prototypes, we only need to learn each annotator's Dirichlet parameters over these prototypes, avoiding to learn a single confusion matrix for each annotator. The advantages are twofold. First, by setting $|S|$ much smaller than the number of annotators, we alleviate the issues of data sparsity and class imbalance that plague existing methods, and largely reduce the computational cost. Second, modeling each annotator as a Dirichlet prior distribution over multiple prototypes enables a richer characterization of annotators.

In summary, the contributions of this study are as follows:
\begin{itemize}
    \item We propose a prototype learning-driven Bayesian classifier combination approach to circumvent the data sparsity and class imbalance issues that plague current confusion-matrix-based methods.
    \item We extend annotator modeling from a single confusion matrix to a Dirichlet prior distribution over a set of prototype confusion matrices.
    \item Experiments on 11 real-world datasets demonstrate that we achieve up to 15\% higher accuracy than state-of-the-art methods in the best case, and about 3\% higher average accuracy with less than 10\% of the computational cost.
\end{itemize}

\noindent
To the best of our knowledge, this is the first work that introduces subtraction into confusion-matrix-based annotator modeling to boost both the effectiveness and dimensionality of annotator modeling.
\section{Related Work}
Depending on how annotators are modeled, existing work can be categorized into single-parameter-based approaches \cite{Demartinietal2012,Bonald17,Lietal2019b,Caoetal2020} and confusion-matrix-based approaches \cite{KimGhahramani2012,Lietal2019a,Songetal2021,Zhangetal24,Chenetal25}.

\textbf{Single-parameter-based Methods} assume that each annotator maintains a consistent accuracy across tasks with different truths, without modeling annotators' confusion behaviors. \citet{Whitehilletal2009} construct the log odds of truth probability as a bilinear function of task difficulty and annotator expertise. To capture the confidence in annotator estimates, \citet{Lietal2014} model annotator errors using the Gaussian distribution and exploit the upper bound of the confidence interval to assess annotator reliability. \citet{Lietal2019b} model annotator expertise as drawn from a Gamma distribution and update it via the expectation-maximization framework. \citet{Caoetal2020} model annotator expertise as drawn from a scaled inverse chi-squared distribution and update it via a gradient descent method. Among single-parameter-based methods, BWA \cite{Lietal2019b} achieves the best generalization performance and has computational efficiency second only to MV. However, due to its strong assumption that annotators have consistent accuracy across all classes, its robustness is weaker than that of confusion-matrix-based approaches.

\textbf{Confusion-matrix-based Methods} learn a confusion matrix for each annotator to distinguish the annotator's labeling behavior under different truths. \citet{DawidSkene1979} propose the DS model, which is the first to capture annotator expertise by confusion matrices. \citet{KimGhahramani2012} extend DS \cite{DawidSkene1979} by incorporating Dirichlet priors. \citet{Venanzietal2014} propose detecting communities among annotators to make annotators within the same community similar. Further, \citet{Chu21b} propose two-layer neural networks to capture the global confusion matrix and the individual confusion matrix, respectively, and select annotator behavior using a Bernoulli variable conditioned on task difficulty and annotator expertise. \citet{Lietal2019a} classify tasks with the same truth into subtypes to capture fine-grained annotator capability. \citet{Ibrahimetal19,Ibrahim21} exploit matrix factorization and annotators' co-occurrences to infer confusion matrices and truths alternately. \citet{Chenetal25} develop a Softmax-Gaussian structure to replace the Dirichlet-Multinomial pattern of \cite{KimGhahramani2012}. \citet{Songetal2021} propose detecting the copying behaviors between annotators to penalize the copiers. Besides, \citet{Wuetal2023} devise a multi-view graph embedding model to learn complex relations among tasks and annotators. \citet{Zhangetal24} propose a coupled confusion correction method to mutually refine the learned confusion matrices. Owing to their broader modeling of annotators, confusion-matrix-based methods generally perform better and demonstrate greater potential \cite{Zhangetal24}. Nevertheless, current confusion-matrix-based methods are sensitive to data sparsity and class imbalance, and are struggling with high computational costs.
\begin{table}[hbtp]
  \renewcommand{\arraystretch}{1.1}
  \setlength{\tabcolsep}{3.3pt}
  \caption{Mathematical notations.}\label{notation}
  \begin{tabular}{lll}
    \noalign{\hrule height 0.9pt}
    \ Symbol &  & Description\\
    \noalign{\hrule height 0.6pt}
    \ $T$ &   & Set of tasks\\
    \ $W$ &   & Set of annotators\\
    \ $K$ &   & Set of classes\\
    \ $S$ &   & Set of Prototypes\\
    \ $t_{i}$ &   & $i$-th task, $i\in\{1, \ldots, |T|\}$\\
    \ $w_{j}$ &   & $j$-th annotator, $j\in\{1, \ldots, |W|\}$\\
    \ $k$ &   & $k$-th class, $k\in\{1, \ldots, |K|\}$\\
    \ $s$ &   & $s$-th prototype, $s\in\{1, \ldots, |S|\}$\\
    \ $z_{i}$ &   & Truth (true annotation) of $t_{i}$\\
    \ $y_{ij}$ &  & Annotation provided by $w_{j}$ for $t_{i}$\\
    \ $x_{ji}$ &  & Prototype of $w_{j}$'s expertise pattern on $t_{i}$\\
    \ $W_{i}$ &  & Set of annotators who have annotated $t_{i}$\\
    \ $N_{j}$ &  & Set of tasks that $w_{j}$ have annotated\\
    \ $\vec{\tau}$ &  & Truth distribution over $K$\\
    \ $\vec{\pi}_{j}$ &  &  $w_{j}$'s expertise distribution over $S$\\
    \ $\vec{v}_{sk}$ &  & $s$'s annotation distribution given $k$ as the truth\\
    \ $\vec{u}$, $\vec{\beta}$, $\vec{a}$ &  & Hyperparameters\\
    \noalign{\hrule height 0.9pt}
  \end{tabular}
\end{table}
\section{Problem Formulation}
Frequently used notations are defined in Table \ref{notation}.

\textbf{Multi-Class Classification Annotations.} $y_{ij}\in K$ denotes the annotation provided by annotator $w_{j}$ for task $t_{i}$, where $K$ is a set of three or more mutually exclusive classes. $z_{i}\in K$ represents the sole truth (true annotation) for $t_{i}$.

\textbf{Truth Inference.} Given the sets of tasks $T$, annotators $W$, and annotations $\{y_{ij}\}$, truth inference focuses on inferring the truth $z_{i}$ for each $t_{i}$. This procedure involves iterative estimates of: (i) the truth probability of each annotation, and (ii) the expertise of each annotator. In single-parameter-based methods, annotator expertise is modeled as a single real value, whereas confusion-matrix-based methods model each annotator as a confusion matrix representing different annotation distributions under different truths.
\section{Method}
\subsection{Motivation}
Existing confusion-matrix-based methods learn a confusion matrix for each annotator. DS \cite{DawidSkene1979} is the most straightforward method, where confusion matrices are estimated via maximum likelihood estimation (MLE). IBCC \cite{KimGhahramani2012}, CBCC \cite{Venanzietal2014}, EBCC \cite{Lietal2019a}, and FGBCC \cite{Chenetal25} extend DS by incorporating prior distributions or by modeling annotators and tasks at a finer granularity. Consider a real-world dataset, Val5, which consists of 100 tasks, 38 annotators, and 5 classes, with a total of 1000 annotations. The truth distribution is $\{0.13, 0.27, 0.23, 0.28, 0.09\}$. The accuracy of Majority Voting (MV), DS, IBCC, CBCC, EBCC, and FGBCC is presented in Table \ref{example}. Clearly, DS is less effective than the others at achieving fine-grained annotator estimation. However, hampered by limited training data and class imbalance, these fine-grained models substantially fall behind DS.

We attribute this to the additive nature of these methods, which extend DS by introducing additional components. Consequently, they are more sensitive to data sparsity and class imbalance, leading to an inaccurate estimate of each annotator's confusion matrix. Furthermore, due to the complexity of tasks and variability in annotator behaviors, a single confusion matrix is often inadequate to capture an annotator's expertise across all tasks.
\begin{table}[htbp]\small
  \renewcommand{\arraystretch}{1.1}
  \setlength{\tabcolsep}{6pt}
    \caption{Accuracy of MV and confusion-matrix-based methods on Val5.}\label{example}
  \begin{tabularx}{\linewidth}{@{}lllllll@{}}
    \noalign{\hrule height 0.9pt}
     \ \ Dataset & MV     & DS   & IBCC & CBCC & EBCC   & FGBCC \ \ \\
     \noalign{\hrule height 0.75pt}
     \ \ Val5    & 0.352 & 0.41 & 0.33 & 0.33 & 0.331  & 0.38 \ \ \\
    \noalign{\hrule height 0.9pt}
  \end{tabularx}
\end{table}
\begin{figure*}[htbp]
\centering
\includegraphics[width=0.84\linewidth]{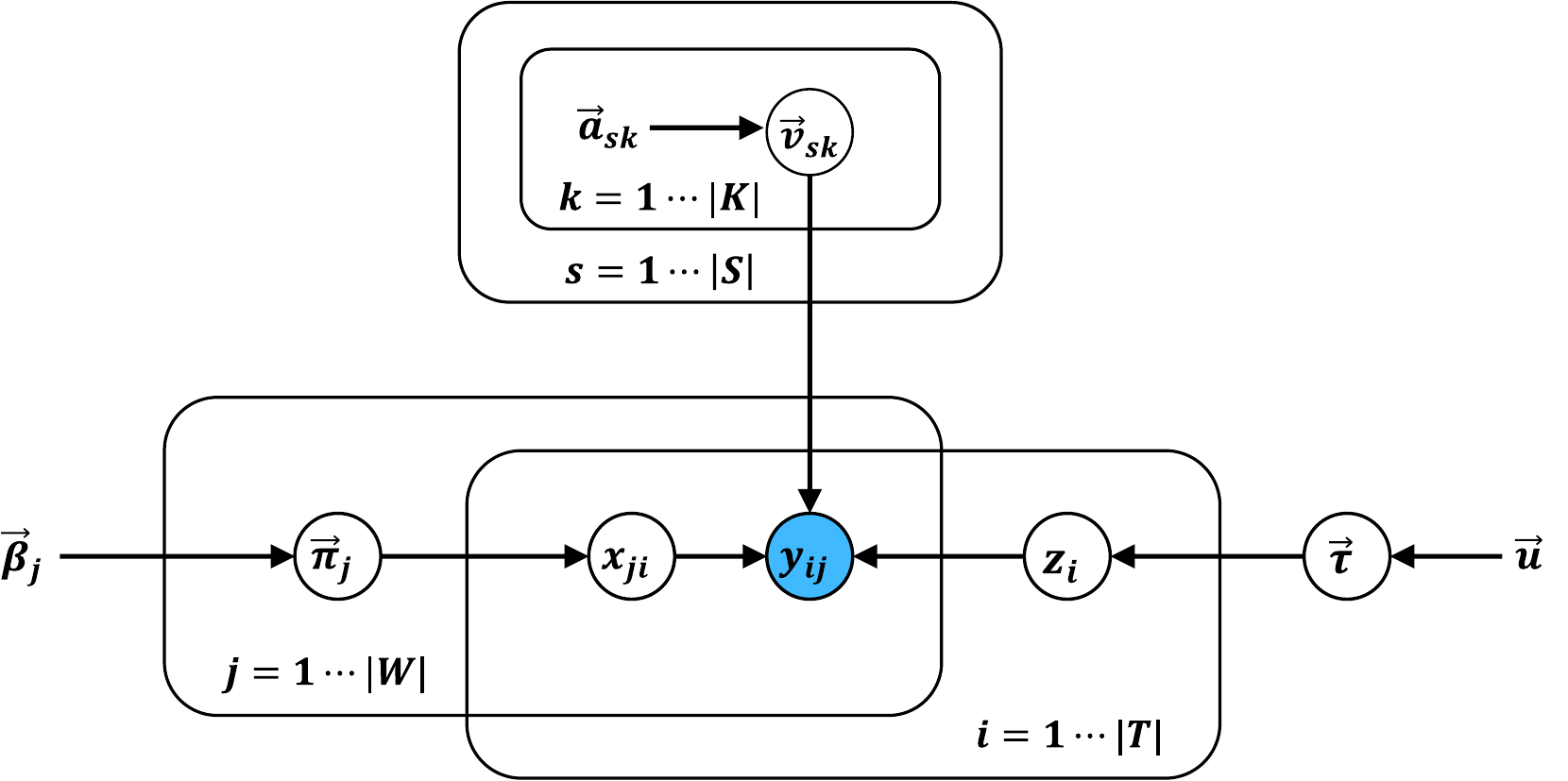}
\caption{The plate notation of PTBCC.}
\label{fig0}
\end{figure*}

To address the issues above, we propose abandoning the traditional means of independently learning a confusion matrix for each annotator. Intuitively, we believe that there exists a set $S$ of prototype confusion matrices, which accommodate the expertise patterns (confusion matrices) of all annotators. Namely, $S$ captures the essential expertise patterns of all annotators. Specifically, we model each annotator as following a Dirichlet prior distribution over these prototypes. Then, $|S|$ prototypes can be reliably trained from all annotators. This prototype learning-driven mechanism not only provides deep insights into the characteristics of all annotators but also circumvents the issues of data sparsity and class imbalance. Furthermore, we can significantly enhance the inference efficiency by setting $|S|\ll|W|$.
\subsection{The Generative Process and Joint Distribution}
Figure \ref{fig0} illustrates the plate notation of PTBCC. We model the truth distribution using a Dirichlet distribution. We assume that there exists a set $S$ of prototype confusion matrices, where each row distribution of $s\in S$ is modeled by a Dirichlet distribution, and the distribution of $w_{j}$'s expertise pattern follows a Dirichlet distribution over $S$. Let $x_{ji}$ denote the prototype index of $w_{j}$'s expertise pattern on $t_{i}$, and $z_{i}$ denote the truth of $t_{i}$. Then, we can derive that the annotation probability of $y_{ij}$ is $v_{x_{ji}z_{i}y_{ij}}$.

The generative process of PTBCC is as follows:

\begin{enumerate}[label=\arabic*., leftmargin=2.3em]
    \item Draw truth distribution:

    $\vec{\tau} \mid \vec{u} \sim \text{Dirichlet}(\vec{u})$
    \item For each annotator $w_{j}$ ($j\in \{1\ldots |W|\}$):
     \begin{itemize}[leftmargin=1.5em]
            \item Draw prototype distribution:

            $\vec{\pi}_{j} \mid \vec{\beta}_{j} \sim \text{Dirichlet}(\vec{\beta}_{j})$
     \end{itemize}
    \item For each prototype $s$ ($s\in \{1\ldots |S|\}$):
    \begin{enumerate}[label*=\arabic*., leftmargin=2em]
    \item Given class $k$ ($k\in \{1\ldots |K|\}$) being the truth:
            \begin{itemize}[leftmargin=1.5em]
            \item Draw annotation distribution:

            $\vec{v}_{sk} \mid \vec{a}_{sk} \sim \text{Dirichlet}(\vec{a}_{sk})$
            \end{itemize}
     \end{enumerate}
    \item For each task $t_{i}$ ($i \in \{1\ldots |T|\}$):
    \begin{enumerate}[label*=\arabic*., leftmargin=2em]
        \item Draw truth assignment:

        $z_i \mid \vec{\tau} \sim \text{Categorical}(\vec{\tau})$
        \item For each annotator $w_{j} \in W_{i}$:
        \begin{enumerate}[label*=\arabic*., leftmargin=2.8em]
        \item Draw prototype assignment:

        $x_{ji} \mid \vec{\pi}_{j} \sim \text{Categorical}(\vec{\pi}_{j})$
        \item Draw annotation assignment:

        $y_{ij} \mid \vec{v}_{x_{ji}z_{i}} \sim \text{Categorical}(\vec{v}_{x_{ji}z_{i}})$
        \end{enumerate}
    \end{enumerate}
\end{enumerate}
Following this generative process, the joint distribution of all observations, latent variables, and parameters, is:
\begin{align}
  &p(Y, Z, V, X, \pi, \tau \mid u, \beta, a) \notag\\
  &=p(\tau\mid u)p(\pi\mid\beta)p(V\mid a)p(Z\mid\tau)p(X\mid\pi)p(Y\mid X,Z,V) \notag\\
  &=\text{Dir}(\vec{\tau} \mid \vec{u})\cdot \prod_{j=1}^{|W|}\text{Dir}(\vec{\pi}_{j} \mid \vec{\beta}_{j})\cdot\prod_{s=1}^{|S|}\prod_{k=1}^{|K|}\text{Dir}(\vec{v}_{sk} \mid\vec{a}_{sk}) \notag\\
  &\cdot\prod_{i=1}^{|T|}\tau_{z_{i}}\prod_{j\in W_{i}}\pi_{jx_{ji}}v_{x_{ji}z_{i}y_{ij}}
  \label{pf1}
\end{align}

The expertise pattern of each annotator is extended from a single confusion matrix to a distribution over multiple prototypes, which are co-trained by all annotators.
\subsection{Inference with Variational Methods}
Since direct inference is computationally intractable, we employ the mean-field approximation in variational inference \cite{Bishop2006,Bleietal2017} to approximate the posterior. The principle is to write the log evidence probability as:
\begin{align}
  &\log p(C)\notag\\
  &=\int q(B)\log{\frac{p(C, B)}{q(B)}}{\rm{d}}Z
  +\int q(B)\log{\frac{q(B)}{p(B|C)}}{\rm{d}}B \notag\\
  &=L(q)+KL(q(B)\|p(B|C))
  \label{pf2}
\end{align}
where $C$ and $B$ denote the set of observed data and latent variables, respectively. Thus, the posterior can be approximated by maximizing $L(q)$, and by applying the mean-field strategy,
\begin{align}
q(B)=\prod_{h=1}^{H}q_{h}(B_{h})
\label{qf1}
\end{align}
we can achieve the optimal $q_{h}^{*}(B_{h})$:
\begin{align}
q_{h}^{*}(B_{h})=\exp(\mathbb{E}_{q(B_{\neg h})}\log p(C, B))
\label{qf2}
\end{align}
Therefore, the goal is to find a variational distribution $q$,
\begin{align}
  &q(Z, V, X, \pi, \tau) \notag\\
  &=\text{Dir}(\vec{\tau} \mid \vec{\nu})\cdot \prod_{j=1}^{|W|}\text{Dir}(\vec{\pi}_{j} \mid \vec{\eta}_{j})\cdot\prod_{s=1}^{|S|}\prod_{k=1}^{|K|}\text{Dir}(\vec{v}_{sk} \mid\vec{\mu}_{sk}) \notag\\
  &\cdot\prod_{i=1}^{|T|}q(z_{i}\mid \vec{\phi}_{i})\prod_{j=1}^{|W|}\prod_{i\in N_{j}}q(x_{ji}\mid \vec{\theta}_{ji})
  \label{qf3}
\end{align}
Based on the variational distribution, Equation (\ref{pf2}), the lower bound of the log evidence probability (Equation (\ref{pf1})) can be written as:
\begin{align}
  &\log L(q)=\sum_{k=1}^{|K|}(u_{k}-\nu_{k}+\sum_{i=1}^{|T|}\phi_{ik})\mathbb{E}_{q}\log\tau_{k}+\log B(\vec{\nu}) \notag\\
  &+\sum_{j=1}^{|W|}\{\sum_{s=1}^{|S|}(\beta_{js}-\eta_{js}+\sum_{i\in N_{j}}\theta_{jis})\mathbb{E}_{q}\log\pi_{js}+\log B(\vec{\eta}_{j})\} \notag\\
  &+\sum_{s=1}^{|S|}\sum_{k=1}^{|K|}\{\sum_{l=1}^{|K|}(a_{skl}-\mu_{skl} \notag\\
  &+\sum_{j=1}^{|W|}\sum_{i\in N_{j}}\phi_{ik}\theta_{jis}\mathbf{1}[y_{ij}=l])\mathbb{E}_{q}\log v_{skl}+\log B(\vec{\mu}_{sk})\}\notag\\
  &-\sum_{i=1}^{|T|}\sum_{k=1}^{|K|}\phi_{ik}\log\phi_{ik}-\sum_{j=1}^{|W|}\sum_{i\in N_{j}}\sum_{s=1}^{|S|}\theta_{jis}\log\theta_{jis}+Const
  \label{qf4}
\end{align}
where $\mathbf{1}[\cdot]$ is an indicator function that returns 1 if $[\cdot]$ is true and 0 otherwise.
\subsection{Parameter Update}
Given the property of the Dirichlet distribution, the expectations of the log parameters are computed as:
\begin{align}
  &\mathbb{E}_{q}\log\tau_{k}=\psi(\nu_{k})-\psi(\sum_{k=1}^{|K|}\nu_{k}) \label{qf5}\\
  &\mathbb{E}_{q}\log\pi_{js}=\psi(\eta_{js})-\psi(\sum_{s=1}^{|S|}\eta_{js}) \label{qf6}\\
  &\mathbb{E}_{q}\log v_{skl}=\psi(\mu_{skl})-\psi(\sum_{l=1}^{|K|}\mu_{skl})
  \label{qf7}
\end{align}

To derive each term in Equation (\ref{qf3}), we sequentially maximize Equation (\ref{qf4}) with respect to each term. Specifically, we update each term based on the current estimates of all the others. Following the standard mean-filed variational Bayes method (Equation (\ref{qf2})), we derive the update rules:
\noindent
\begin{align}
  &\nu_{k}=u_{k}+\sum_{i=1}^{|T|}\phi_{ik}  \label{qf8}\\
  &\eta_{js}=\beta_{js}+\sum_{i\in N_{j}}\theta_{jis}  \label{qf9}\\
  &\mu_{skl}=a_{skl}+\sum_{j=1}^{|W|}\sum_{i\in N_{j}}\phi_{ik}\theta_{jis}\mathbf{1}[y_{ij}=l] \label{qf10}\\
  &\theta_{jis}\propto \exp(\mathbb{E}_{q}\log\pi_{js}+\sum_{k=1}^{K}\phi_{ik}\mathbb{E}_{q}\log v_{sky_{ij}}) \label{qf11}
\end{align}
\begin{equation}
  \phi_{ik}\propto \exp(\mathbb{E}_{q}\log\tau_{k}+\sum_{j\in W_{i}}\sum_{s=1}^{|S|}\theta_{jis}\mathbb{E}_{q}\log v_{sky_{ij}})
  \label{qf12}
\end{equation}
Equations (\ref{qf5}) -- (\ref{qf12}) are iteratively updated until the change in each $\phi_{ik}$ is smaller than $\xi$.
\subsection{Generality of Our Method}
Data sparsity and class imbalance are common challenges in machine learning. Our prototype learning-driven mechanism not only ensures the reliable training of prototype confusion matrices but also extends annotator expertise modeling from a single confusion matrix to a distribution over multiple prototype confusion matrices. By centralizing data for prototype learning, our method offers critical insights for other fields of machine learning.
\begin{figure*}[htbp]
\centering
\includegraphics[width=0.97\linewidth]{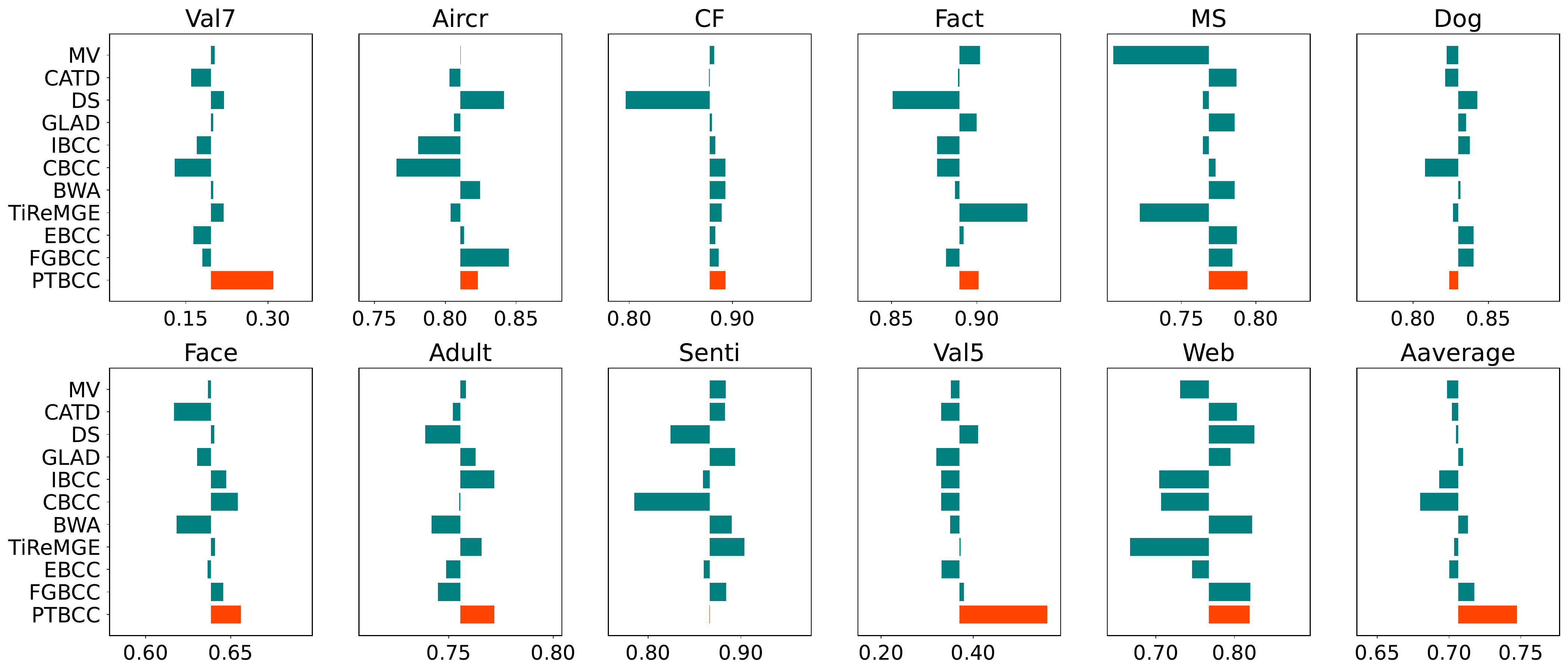}
\caption{Comparison of accuracy on 11 datasets, with a horizontal baseline indicating the average accuracy.}
\label{Acc1}
\end{figure*}
\begin{table}[h]\small
\centering
\renewcommand{\arraystretch}{1.25}
    \setlength{\tabcolsep}{2.1pt}
        \caption{Statistics of 11 real-world datasets.} \label{dataset}
    \begin{tabularx}{0.989\linewidth}{@{}lrrrclr@{}}
    \noalign{\hrule height 1pt}
     \ \ Dataset & \#Task & \#Annotator & \#Truth & \#Class & \#Labels & Domain \ \  \\
    \noalign{\hrule height 0.7pt}
     \ \ Val7 & 100 & 38 & 100 & 7 & 1000   & Sentiment \ \   \\
     \ \ Aircr & 593 & 50 & 593 & 6 & 1588   & Image \ \  \\
     \ \ CF & 300 & 461 & 300 & 5 & 1720   & Sentiment \ \  \\
     \ \ Fact & 42,624 & 57 & 576 & 3 & 214,915 &   text  \ \  \\
     \ \ MS & 700 & 44 & 700 & 10 & 2945   & Audio \ \   \\
     \ \ Dog & 807 & 109 & 807 & 4 & 8070   & Image \ \   \\
     \ \ Face & 584 & 27 & 584 & 4 & 5242   & Image \ \   \\
     \ \ Adult & 11,040 & 825 & 333 & 4 & 89,799   & Level \ \   \\
     \ \ Senti & 98,980 & 1960 & 1000 & 5 & 569,274  & Sentiment \ \   \\
     \ \ Val5 & 100 & 38 & 100 & 5 & 1000   & Sentiment \ \   \\
     \ \ Web & 2665 & 177 & 2653 & 5 & 15,567   & Relevance \ \   \\
    \noalign{\hrule height 1pt}
    \end{tabularx}
\end{table}
\section{Experiments}
\subsection{Experimental Setup}
\textbf{Datasets.} We employ 11 widely used real-world datasets for evaluation. Eight (CF, Fact, MS, Dog, Face, Adult, Senti, Web) of them are also used in \citet{Lietal2019b}, and the remaining three (Val5, Val7, Aircr) are also used in \citet{Wuetal2022}. Table \ref{dataset} summarizes their statistics.

\noindent
\textbf{Metrics.} Accuracy is the evaluation metric.

\noindent
\textbf{Initialisation.} $\xi=0.001$. We initialize $\phi_{ik}$ by majority voting. We set $|S|=2$, $e=1$, $f=5$, $m=1.35$, with the (diagonal, off-diagonal) of $v_{1}$, $v_{2}$ being ($\frac{f}{f+(|K|-1)\cdot e}, \frac{e}{f+(|K|-1)\cdot e}$), and ($\frac{e}{e+(|K|-1)\cdot m}, \frac{m}{e+(|K|-1)\cdot m}$) to encode our initial belief for two varied patterns. Then,
\begin{align}
  &u_{k}=\sum_{i=1}^{|T|}\phi_{ik};\quad\theta_{jis}=\sum_{k=1}^{K}\phi_{ik}v_{skl}\mathbf{1}[y_{ij}=l] \notag\\
  &\beta_{js}=0.4\sum_{i=1}^{|T|}\theta_{jis};\quad a_{skl}=0.5\sum_{j=1}^{|W|}\sum_{i=1}^{|T|}\theta_{jis}\phi_{ik}\mathbf{1}[y_{ij}=l] \notag\\
  &\theta_{jis}=\frac{\theta_{jis}}{\sum_{s=1}^{|S|}\theta_{jis}} \notag
\end{align}
\textbf{Baselines.} \citet{Zhengetal2017} release their survey implementations, from which we select six methods: MV, DS \cite{DawidSkene1979}, GLAD \cite{Whitehilletal2009}, IBCC \cite{KimGhahramani2012}, and CBCC \cite{Venanzietal2014}, and CATD \cite{Lietal2014}. We also append four recent methods: BWA \cite{Lietal2019b}, EBCC \cite{Lietal2019a}, TiReMGE \cite{Wuetal2023}, and FGBCC \cite{Chenetal25}.

We run all experiments on a server with 2vCPU Intel Xeon Platinum 8369HC, and 8 GB memory.
\subsection{Effectiveness}
Table \ref{Accuracy} and Figure \ref{Acc1} present comparison results of all methods. In Table \ref{Accuracy}, $S$ denotes the sum of ranks where the method is inferior to MV, and the p-value denotes the significance level. We categorize significance levels into 0.1(*) and 0.01(**).
\begin{table}[h]\small
    \centering
    \renewcommand{\arraystretch}{1.25}
    \setlength{\tabcolsep}{8.3pt}
        \caption{Comparison in average accuracy and one-sided Wilcoxon signed-rank test against MV.} \label{Accuracy}
    \begin{tabularx}{0.99\linewidth}{@{}lcccc@{}}
        \noalign{\hrule height 1pt}
        \ \ \ Method & Avg. Accuracy & $S$ & Sig.level & p--value  \\
        \noalign{\hrule height 0.7pt}
        \  \ \ MV          &  0.6986 &   &   &  \ \  \\
        \  \ \ CATD        &  0.7020  &  45 & & 0.8608 \ \  \\
        \  \ \ DS       &  0.7049   & 27  &  & 0.3188 \ \  \\
        \  \ \ GLAD       &  0.7099 & 26 &   & 0.2886 \ \  \\
        \  \ \ IBCC        &  0.6932  & 45 &   & 0.8608 \ \  \\
        \  \ \ CBCC         &  0.6798   &  51 &   & 0.9492 \ \  \\
        \  \ \ BWA        &  \textbf{0.7132}   & 27 &   & 0.3188 \ \  \\
        \  \ \ TiReMGE         &  0.7036 &   14 &   *  & 0.0508 \ \  \\
        \  \ \ EBCC       &  0.7002  &  37 &    &  0.6499\ \  \\
        \  \ \ FGBCC      &  \textbf{0.7175} &  17 &  * & 0.0874 \ \  \\
        \noalign{\hrule height 0.7pt}
        \  \ \ PTBCC        &  \textbf{0.7472} &    7  &  ** & 0.0093 \ \  \\
        \noalign{\hrule height 1pt}
    \end{tabularx}
\end{table}

Table \ref{Accuracy} demonstrates that PTBCC achieves the highest average accuracy, while maintaining the highest significance level in the one-sided Wilcoxon signed-rank test against MV. FGBCC follows; however, due to its complex inference process, it is less robust than PTBCC across all datasets.

BWA is the best-performing single-parameter-based method. Nevertheless, as Figure \ref{Acc1} shows, its performance on Face and Adult lags significantly behind that of MV, resulting in a non-significant outcome in the one-sided Wilcoxon signed-rank test against MV. We attribute this to the greater diversity in annotator expertise in these datasets, which exacerbates the inability of single-parameter methods to capture annotators' confusing behaviors across classes.

Notably, TiReMGE surpasses many baselines in the one-sided Wilcoxon signed-rank test against MV. From Figure \ref{Acc1}, we find that TiReMGE consistently outperforms MV owing to its multi-perspective modeling of annotators and tasks.

Furthermore, on the two largest datasets, Senti and Fact, PTBCC keeps a clear advantage over most confusion-matrix-based methods. Moreover, PTBCC achieves a slightly below-average accuracy on Dog. We examined the results and found that the prototypes on this dataset exhibit notably accurate annotating behaviors and less distinctiveness, which can be attributed to the overall high quality of annotators and abundant training data.

Overall, our prototype learning-driven mechanism effectively overcomes the issues of limited training data and class imbalance, while extending the modeling of annotator expertise from a single confusion matrix to a Dirichlet distribution over prototype confusion matrices.
\subsection{Case Studies on Val5 and MS}
\begin{figure*}[htbp]
\centering
\includegraphics[width=0.95\linewidth]{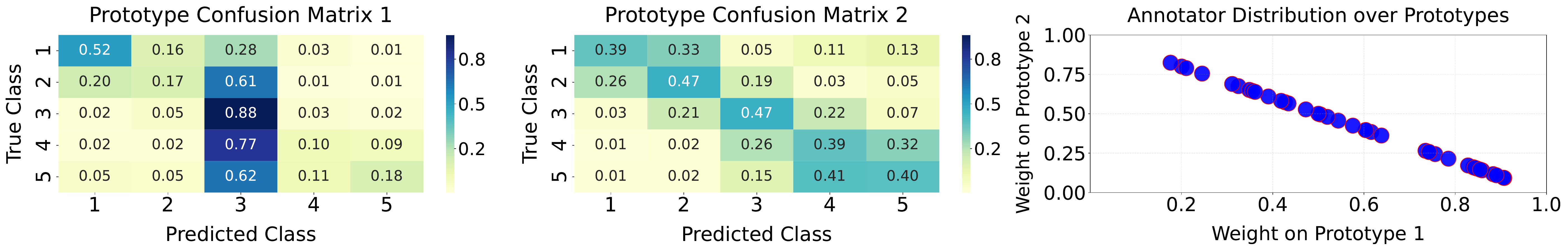}
\caption{Prototype confusion matrices and annotator distributions learned from Val5.}
\label{Overall-Val5}
\end{figure*}

\begin{figure*}[htbp]
\centering
\includegraphics[width=0.95\linewidth]{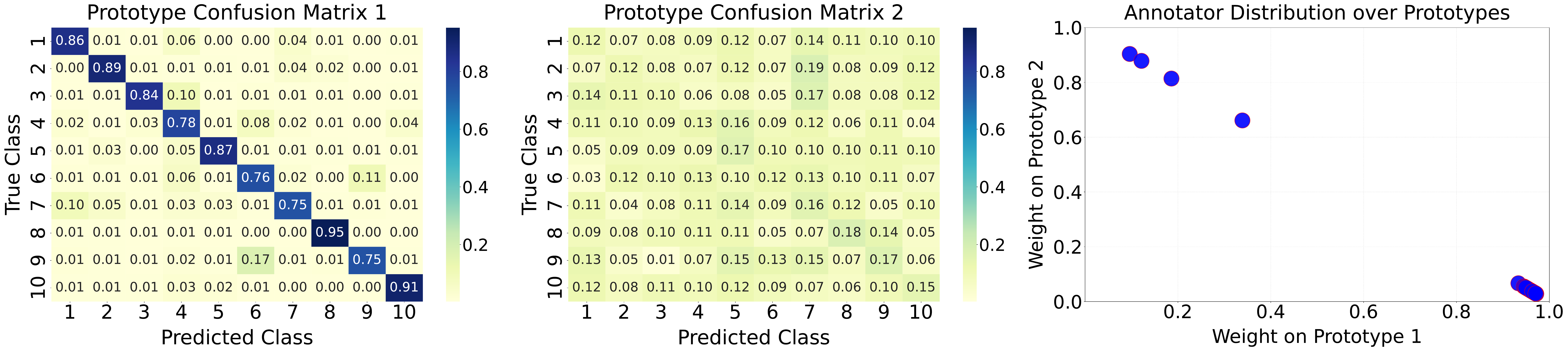}
\caption{Prototype confusion matrices and annotator distributions learned from MS.}
\label{Overall-MS}
\end{figure*}

Figure \ref{Overall-Val5} and Figure \ref{Overall-MS} present the prototype confusion matrices and corresponding annotator distributions learned on Val5 and MS, respectively. It is evident that the prototypes are highly distinctive in both datasets.

From Figure \ref{Overall-Val5}, we are easy to find that the annotators in Val5 are generally of low expertise. Prototype 1 frequently confuses other classes with class 3, especially misclassifying class 4 as class 3, while maintaining relatively higher accuracy when the truth is class 1. In contrast, Prototype 2 demonstrates a higher accuracy, with the annotation probability for the correct class much higher than for other classes. Interestingly, both prototypes reveal a tendency to confuse class 1 with class 2, and class 4 with class 5, indicating the essential identity of this dataset. All annotators are roughly uniformly distributed but slightly skewed towards Prototype 1, indicating that the tasks may be inherently difficult. The low-quality annotations also explain the poor performance of all methods.

Figure \ref{Overall-MS} shows significant variation in the prototype confusion matrices in MS, with most annotators favoring Prototype 1, which demonstrates high classification accuracy across all classes. In contrast, Prototype 2 denotes the random classification patterns. This helps understand why all methods achieve notably higher accuracy on MS.

We demonstrate remarkable performance on Val5 and MS, with a 15\% accuracy improvement on Val5. Furthermore, we offer valuable insights into annotator structures.
\subsection{Ablation Study}
Table \ref{abl} presents the average accuracy of PTBCC with respect to varied $f$, $m$, and $|S|$. In addition to the initial two prototypes, we add prototypes generated from a uniform Dirichlet distribution. 3-Ran denotes a confusion matrix where all entries equal to $\frac{1}{|K|}$.

\noindent
\textbf{Choosing $f$ and $m$.} From Table \ref{Accuracy}, it can be observed that under varied $f$ or $m$, PTBCC consistently outperforms state-of-the-art methods, demonstrating the reliability of training two prototypes. Moreover, within a reasonable range, initial prototypes with greater differentiability yield better training performance.
\begin{table}[htbp]\small
  \renewcommand{\arraystretch}{1.1}
  \setlength{\tabcolsep}{2.8pt}
    \caption{Average accuracy of PTBCC with varied $f$, $m$, and $|S|$, respectively.}\label{abl}
  \begin{tabularx}{\linewidth}{@{}l|lllllll@{}}
    \noalign{\hrule height 0.9pt}
     \ \ $f$              & 2         &  3      &      4    & 5       & 6       & 7         & 8  \ \  \\
     \ \ Accu       & 0.7358    & 0.7443  &    0.7447 & 0.7472  & 0.7469  & 0.7464    &   0.7467 \ \ \\
    \noalign{\hrule height 0.5pt}
    \ \ $m$                &  1.1      & 1.15      &  1.2       &   1.25    & 1.3       & 1.35      & 1.4   \ \    \\
     \ \ Accu              &  0.7445   &   0.7439  &  0.7447    &  0.7447   & 0.7467    &  0.7472   &  0.7461 \ \  \\
    \noalign{\hrule height 0.5pt}
    \ \ $|S|$       & 2            &  3-Ran &  3         &  4      & 5      &  6      &   7   \ \     \\
     \ \ Accu      &  0.7472      &  0.7288   &  0.7300      &  0.7271 & 0.7259 & 0.7229  & 0.7205  \ \   \\
    \noalign{\hrule height 0.9pt}
  \end{tabularx}
\end{table}

\noindent
\textbf{Number of Prototypes.} As shown in Table \ref{abl}, the average accuracy of PTBCC slowly decreases as the number of prototypes increases. We attribute this to the fact that, with more prototypes, the distribution of annotators over prototypes becomes sparser, leading to lower confidence in the prototype confusion matrices. Nevertheless, PTBCC still surpasses the baselines.
\subsection{Efficiency}
\begin{figure}[htbp]
\centering
\includegraphics[width=\linewidth]{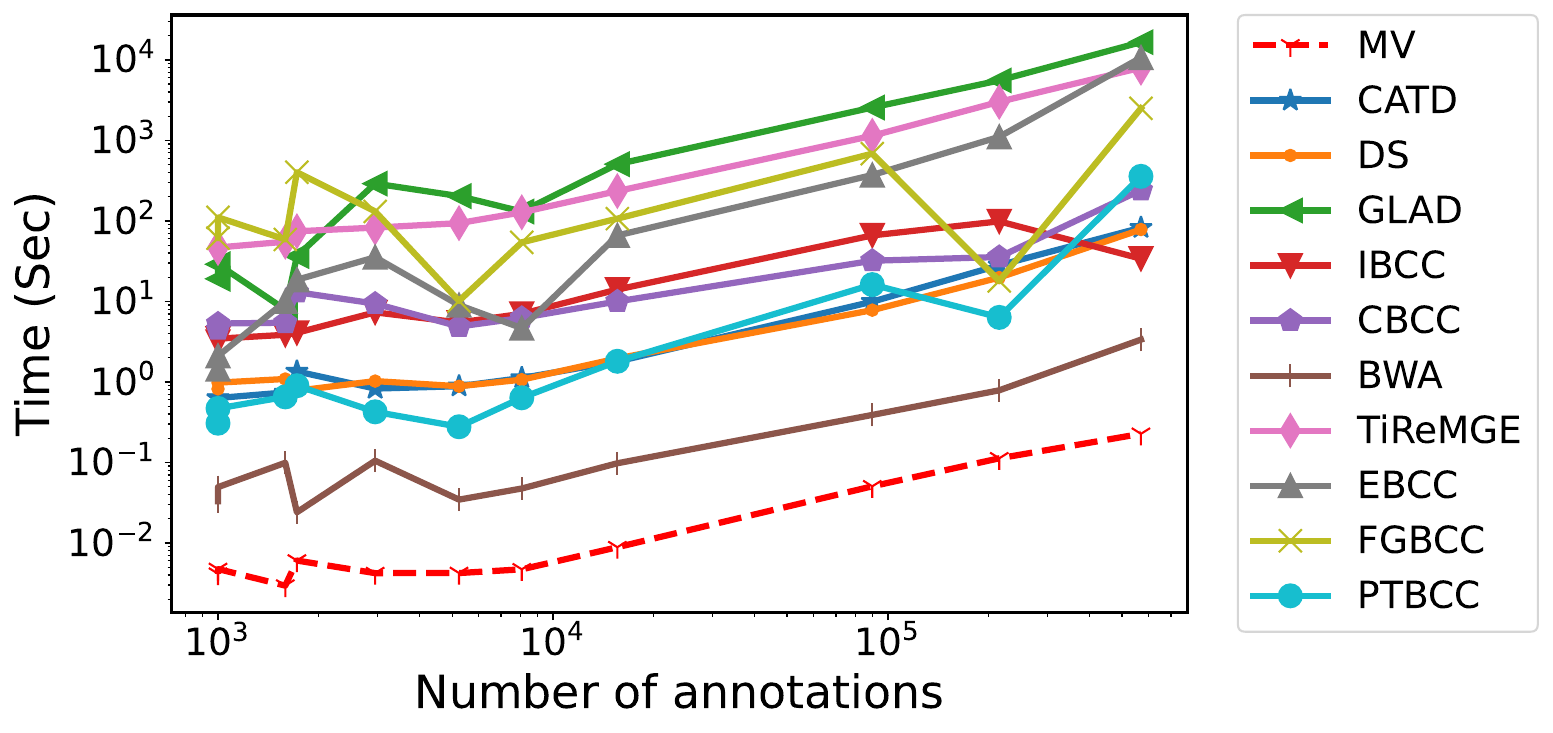}
\caption{Comparison of running time on 11 datasets.}
\label{runtime}
\end{figure}

Figure \ref{runtime} compares the runtime on 11 datasets arranged in ascending order of the annotation size. The runtime of all methods generally increases with the number of annotations. MV, as the most naive method, consistently ranks first. BWA achieves the highest performance among single-parameter-based methods, following MV. Our method ranks third most of the time, holding a clear superiority over confusion-matrix-based methods.

DS updates via maximum likelihood estimation, exhibiting greater efficiency than EBCC and FGBCC, which involve fine-grained inference. GLAD exhibits the lowest efficiency; we attribute this to its incorporation of task difficulty and the gradient-based inference. TiReMGE requires a long training time, mainly due to the optimization of an objective function that captures complex relationships among annotators and tasks.

Overall, confusion-matrix-based methods generally require much longer training time. Our method demonstrates outstanding performance in both effectiveness and efficiency, achieving an average accuracy improvement of 3\% with less than 10\% of the running time.
\section{Conclusion}
This paper addresses the challenges of data sparsity and class imbalance faced by existing confusion-matrix-based methods. Our prototype learning-driven mechanism extends annotator modeling from a single confusion matrix to a distribution over multiple prototype confusion matrices. We explore the prototypes that capture the inherently complex annotator expertise across all tasks. We achieve reliable prototype training and richer annotator estimation by projecting each annotator onto a Dirichlet prior distribution over prototypes. Notably, our work is the first to apply subtraction operations to confusion-matrix-based methods for enhanced annotator estimation. Extensive experiments demonstrate both the effectiveness and efficiency of our method.
\bibliography{aaai2026}


\end{document}